# Fuzzy Distribution Modeling for Synthetic Tabular Data Generation with Causality Preservation

Michael Vasilakakis, Dimitris K. Iakovidis
Department of Computer Science and Biomedical Informatics
University of Thessaly
Lamia, Greece
{vasilaka, diakovidis}@uth.gr

***Abstract*— Synthetic tabular data generation provides an effective alternative for the training of machine learning models when real-world data is limited or inaccessible. However, the heterogeneous, non-smooth, and incomplete nature of tabular data poses fundamental challenges to conventional probabilistic and deep generative models, where their interpretability remains limited. This paper proposes a novel fuzzy distribution modeling methodology for synthetic tabular data generation based on fuzzy sets theory. Feature distributions are represented using fuzzy sets and feature dependencies are modeled through Fuzzy Cognitive Maps, resulting in a low-parameter, and an interpretable data representation. Synthetic samples are generated by sampling fuzzy concepts rather than raw values, enabling native support for mixed data types, missing values, and domain constraints. The methodology further supports linguistic queries and IF–THEN reasoning, facilitating transparent simulation of decision-making processes. Experimental results on benchmark datasets demonstrate competitive performance with respect to utility, fidelity and privacy compared to state-of-the-art methods, while offering substantially improved interpretability. These results establish fuzzy distribution modeling as a principled and effective approach for synthetic tabular data generation in fuzzy systems and decision support applications.**

***Keywords—Data generation, synthetic tabular data, fuzzy cognitive maps, causality, interpretability***

## I. Introduction

Synthetic data generation (SDG) is a research field focused on creating artificial datasets whose characteristics closely resemble those of the original data, both at the individual sample level and at the population level [1]. The demand for high-quality synthetic tabular data has grown rapidly with the proliferation of data-driven applications across diverse domains. Tabular data is among the most prevalent data types and is widely used in applications such as medical diagnosis, financial analysis, fraud detection, and recommendation systems [2], [3]. Synthetic data has been extensively explored for privacy-preserving machine learning, augmentation of imbalanced datasets, and data sharing under regulatory and legal constraints. By substituting original datasets with synthetic counterparts, sensitive information can be protected, such as personal or customer data, while still enabling downstream analysis and model development [4]. In scenarios where access to real-world data is limited, restricted, or insufficient, synthetic data has emerged as a viable alternative. The primary objective of SDG is to generate training data that faithfully reflects the underlying distributions and dependencies of real data, such that models trained on synthetic data can be effectively deployed on real-world datasets[5].

Existing approaches to synthetic data generation can be broadly categorized into neural and non-neural methods. Neural Generative Adversarial Networks (GANs) and Variational Autoencoders (VAEs) [6] are the two most commonly used architectures. Early VAE-based models for tabular data rely on one-hot encoding for categorical variables and mixture-based normalization schemes for continuous features, combined with simple encoder–decoder architectures[7]. GAN-based models remain the most widely adopted for tabular data synthesis by prepossessing the data features to make their distribution easier to be modeled with conditional tabular GAN [7].

Beyond neural approaches, non-neural methods remain relevant due to their interpretability and statistical grounding. Copula-based models [9], [10] explicitly separate marginal distributions from dependency structures and have been widely applied to tabular data generation. However, parametric copulas often struggle in high-dimensional settings [7]. Probabilistic graphical models [9] offer an alternative by directly modeling variable dependencies and have shown effectiveness in privacy-preserving data synthesis, though they typically require prior knowledge of the dependency graph or large sample sizes for reliable structure learning. The tree ensembles exhibit good performance on tabular prediction tasks [11]. Several works have proposed tree-based generative models, including adversarial and diffusion-inspired forests; however, while these methods can be effective, scalability issues arise [12].

Generating high-quality synthetic tabular data presents several technical challenges that are not encountered in text or image generation [6]. Tabular data are characterized by heterogeneous feature types, which means that their features can take continuous, categorical, and discrete values. Tabular data are characterized by distributions that are non-smooth and imbalanced. In addition, access to high-quality tabular datasets is frequently limited because of incomplete records, privacy constraints, or the high cost of data collection, further motivating the need for effective synthetic data generation. Another issue of tabular data is their diversity, because they vary widely in the number of features, feature composition and value ranges. As a result, no single generative approach can be expected to perform consistently well across all tabular domains [1]. Thus, due to this diversity, there is no universal embedding representation for tabular data, such as the word embedding considered in natural language processing or deep feature extractors in computer vision. This lack of transferable representations poses significant challenges for tabular SDG, particularly when learning from small, noisy, or skewed datasets [6]. Furthermore, tabular data often contain complex and domain-specific dependencies between features, while correlations among columns are typically weaker than the

This work is part of the European project SEARCH, which is supported by the Innovative Health Initiative Joint Undertaking (IHI JU) under grant agreement No. 101172997. The JU receives support from the European Union's Horizon Europe research and innovation programme and COCIR, EFPIA, Europa Bio, MedTech Europe, Vaccines Europe, Medical Values GmbH, Corsano Health BV, Syntheticus AG, Maggioli SpA, Motilent Ltd, Ubitech Ltd, Hemex Benelux, Hellenic Healthcare Group, German Oncology Center, Byte Solutions Unlimited, AdaptIT GmbH. Views and opinions expressed are however those of the author(s) only and do not necessarily reflect those of the aforementioned parties. Neither of the aforementioned parties can be held responsible for them.

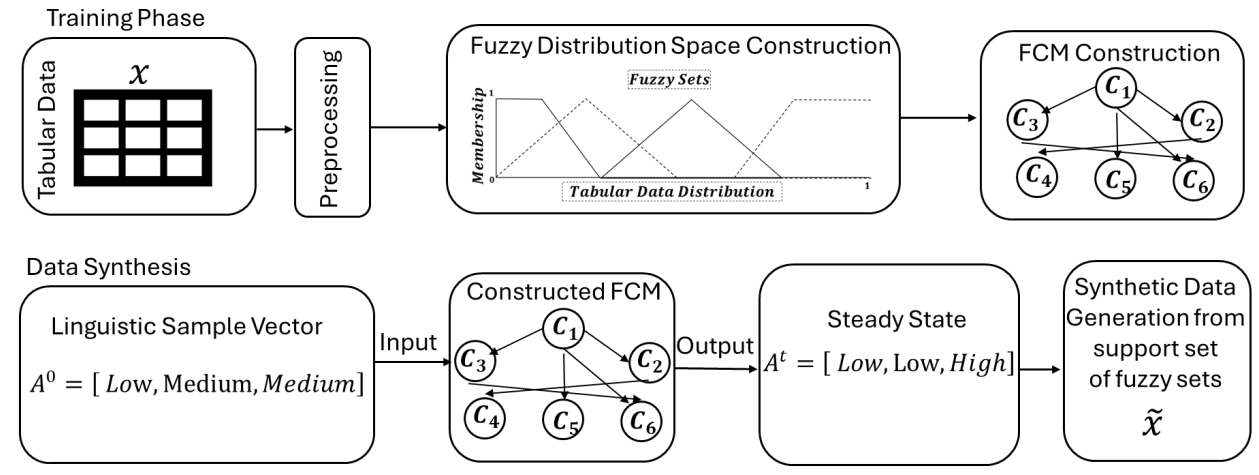

Fig. 1. Overview of the proposed synthetic tabular data generation based on fuzzy distribution space and FCM

spatial correlations in images or the semantic relationships in language data. This is because tabular features do not possess inherent positional structure [1]. Furthermore, tabular data are usually well-defined and semantically grounded, with each feature corresponding to a meaningful real-world quantity, such as age, income, or medical measurements. This observation motivates the modeling of feature interactions to describe the causal relationships among them [2].

Causality describes how variables influence one another to give rise to observed data and offers a more informative representation than correlation alone. According to the common cause principle, every observed correlation is either the result of a direct causal relationship or is induced by a shared underlying factor [2]. Incorporating causal structure into synthetic data generation therefore holds the potential to improve the fidelity, consistency, and interpretability of synthetic tabular data.

The theoretical framework of fuzzy sets provides a mathematically sound foundation for the development of systems that are robust to uncertainty and imprecision inherent in real-world data [13]. Fuzzy classification systems, using linguistic variables and membership functions, incorporate semantic knowledge, that is directly interpretable by humans, enabling transparent reasoning and explainable decision processes. Beyond classification tasks, fuzzy logic has been successfully applied to address complex scientific and engineering problems where uncertainty, heterogeneity, and non-linear interactions are present. Among these approaches, Fuzzy Cognitive Maps (FCMs) stand out as graph knowledge-based methods, defining concepts and their causal relationships [14]. By extending fuzzy logic with causal reasoning, FCMs offer an effective mechanism for modeling interdependencies among tabular data features and capturing complex system dynamics [15].

To address the challenges of synthetic tabular data generation, this paper proposes a fuzzy distribution modeling framework for synthetic tabular data generation. The proposed approach exploits fuzzy membership functions to represent feature marginals and fuzzy relations to characterize inter-feature dependencies, enabling a joint fuzzy representation of the data structure. In contrast to conventional probabilistic and neural generative models that rely on precise distributional assumptions or latent embeddings, the proposed framework explicitly accounts for vagueness, imprecision, and semantic meaning in tabular data. Synthetic samples are generated by sampling fuzzy concepts rather than raw feature values, naturally supporting mixed data types, missing values, and domain constraints, while enabling causal reasoning through FCMs. To the best of our knowledge, this work constitutes one of the first attempts to investigate fuzzy distribution modeling as a systematic approach for synthetic tabular data generation, offering an interpretable and computationally efficient alternative for decision-support and data-driven applications.

The rest of the paper is organized as follows. Section II presents the proposed fuzzy distribution modeling framework. Section III reports experimental results and comparisons with state-of-the-art methods. Section IV concludes the paper and outlines future research directions.

## II. The Proposed Methodology

### *A. Data Preprocessing and Normalization*

The first step of the methodology involves appropriate data preprocessing and normalization (Fig.1). Let $D = \{(x_m, y_m) | x_m \in \mathbb{R}^d, y_m = 1, \dots, K,\ m = 1, \dots, M\}$ , denote the original label set of training tabular data, where $x_m$ is the $m$-th feature vector and $y_m$ is the corresponding class label for a $K$ class dataset. To obtain normalized representations suitable for fuzzy distribution modeling, each feature is first standardized using z-score normalization based on its empirical mean and standard deviation, reducing scale bias and improving numerical stability in subsequent clustering steps. The intermediate z-score step reduces extreme inter-feature scale differences and improves numerical stability in the subsequent k-means clustering used to define fuzzy prototypes. The standardized features are then linearly rescaled to the interval [0,1], defining a common universe of discourse across all variables. The resulting normalized dataset is denoted as $D' = \{(x'_m, y_m)\}$. This normalization facilitates the definition of fuzzy membership functions and enables consistent modeling of heterogeneous features within a unified fuzzy framework.

### *B. Fuzzy Distribution Construction*

The proposed methodology reformulates synthetic tabular data generation as a fuzzy causal reasoning problem rather than a density estimation task. The second step of the methodology involves constructing appropriate fuzzy sets for each feature of the feature vectors belonging to each class $k = 1 \dots K$ independently. These fuzzy sets serve as the basis for the characterization of initial features and the construction of the fuzzy distribution (Fig.1). Let $D'_k = \{(x'_m, y_m) | y_m = k\}$, denote the subset of samples belonging to class $k = 1 \dots K$. For the features $x'_{m,f}, f = 1, \dots, d$ of a feature vector $x'_m$ belonging to class $k$, the values are clustered using a clustering algorithm, such as k-means, to identify representative values they take within class $k$. Let $V_f^k = \{(v_{e,f}^k, y_{c,f}) | y_{c,f} = k\}$ denote the resulting cluster centers, where $v_{e,f}^k$ is the value of the center of the cluster $e = 1, \dots, E$, formed by the features $x'_{m,f}$. $E$ corresponds to the number of linguistic values, which will be used for the fuzzification of the features $x'_{m,f}$, as described in the following, *e.g.,* for $E = 3$ the linguistic values can be "Low", "Medium", and "High". This feature-wise clustering process aims to capture the dominant value regions of each feature.

The fuzzification process considers that for each component $v_{e,f}^k, y_{e,f} = k$ a corresponding fuzzy set $A_{e,f}^k, y_{e,f} = k$ is defined, representing the linguistic concept associated with the $e$-th cluster. This set is characterized by a membership function $\mu_{e,f}^{\text{k}}(x'_{m,f}) \in [0,1]$ , which for simplicity is considered triangular. The top of this triangular function is positioned at $v_{e,f}^k$ while its support extends to the nearest neighboring component $v_{e',f}^k, e \neq e'$ or the boundary

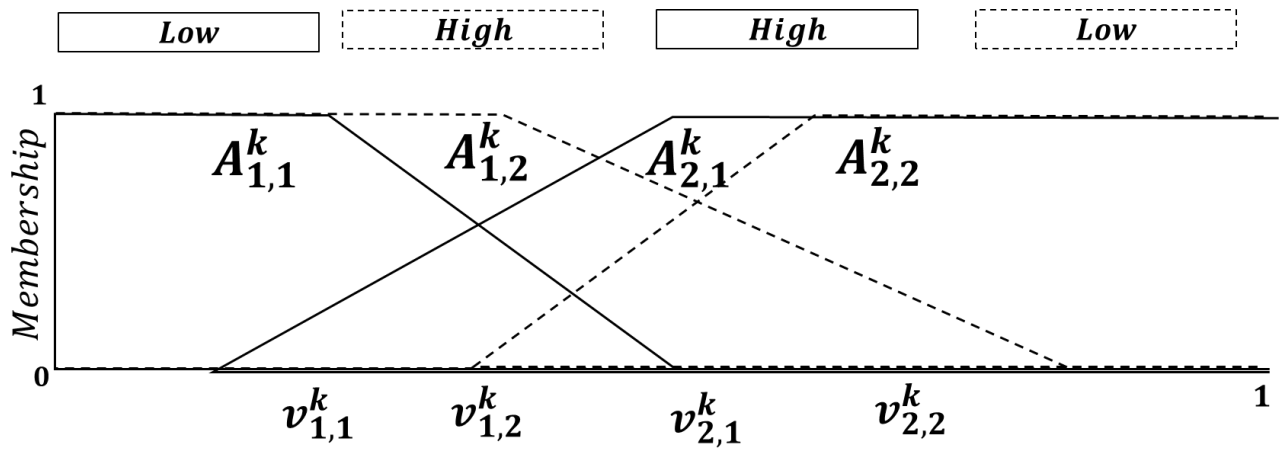

Fig. 2. Illustration of defined fuzzy distribution space, and membership functions of the respective fuzzy sets. Solid lines correspond to the distribution of features $x'_{n,1}$ and the dashed lines correspond to the distribution of features $x'_{n,2}$. Solid and dashed lines distinguish two features of the same class, each simultaneously activating multiple concepts to varying degrees.

values 0 or 1 using trapezoidal membership function, depending on which is closest.

Although clustering identifies representative value regions, the semantic relevance of the corresponding linguistic concepts must be quantified. The linguistic semantics are derived from the degree of each fuzzy set receives from samples $x'_{m,f}$ of the corresponding class. For each fuzzy set $A^k_{e,f}$ the fuzzy cardinality [16] is computed by summing the membership degrees of all elements in the fuzzy set, providing a measure of its effective size :

$$card\left(A^k_{e,f}\right) = \sum_{\forall \mu^k_{e,f}(x'_{m,f})>0} \mu^k_{e,f}\left(x'_{m,f}\right) \quad (1)$$

which measures the total degree to which feature values of class $k$ are compatible with the fuzzy concept represented by $A^k_{e,f}$. The fuzzy cardinality reflects the activity level and representativeness of a fuzzy set within a given class, providing a scalar index of the concept's representativeness within that class. The fuzzy sets are ordered in ascending order so that $card\left(A^k_{1,f}\right) \leq \cdots \leq card\left(A^k_{e,f}\right) \leq \cdots card\left(A^k_{E,f}\right)$ for $e = 1, \ldots, E$. This formulation ensures that strongly expressed linguistic concepts (e.g., *High*) are capable of fully activating the corresponding feature representation, whereas weak linguistic expressions (e.g., *Low*) produce limited activation.

An illustrative example of the definition of fuzzy sets $A_{e,f}$ is presented in Fig.2, where the number of clusters is $E = 2$ and the number of presented features is two $x'_{m,f}, f = 1,2$ of the same class $y_{e,f} = k$. In this figure, horizontal axis presents the normalized feature values $x'_{e,1}$ and $x'_{e,2}$ , while the values $v_{1,1}$ , $v_{1,2}$, $v_{2,1}$ , $v_{2,2}$ indicate the corresponding cluster centers. The fuzzy sets $A_{1,1}$and $A_{1,2}$with solid line represent describe the distribution of features $x'_{m,1}$ and the fuzzy sets $A_{2,1}$and $A_{2,2}$with the dashed line describe the distribution of features $x'_{m,2}$. Vertical axis presents the membership degree of a feature value to belong to a fuzzy set. Fig. 2 illustrates that a single data point may simultaneously activate membership functions of different features to different degrees. This multi-concept co-activation forms the basis for FCM-mediated inter-feature influence propagation during synthesis.

A key advantage of the proposed fuzzy representation is that explicit imputation of missing feature values is not required. When a feature value $x'_{m,f}$ is missing, no crisp membership assignment is computed, and the feature remains partially uninstantiated in the fuzzy space. This preserves uncertainty rather than introducing potentially biased estimates. Missing values are then naturally accommodated through the propagation of influence from observed features during fuzzy cognitive reasoning.

### *C. Modeling Feature Dependencies Using Fuzzy Cognitive Maps*

To capture dependencies and causal interactions among tabular features, the proposed framework constructs a FCM [14] as a causal modeling layer built on top of the fuzzy distribution of the previous subsection. FCMs provide a transparent and interpretable mechanism to represent how feature concepts influence one another, while naturally accommodating uncertainty.

An FCM is defined as a directed weighted graph comprising $N$ concepts $C_i, i = 1,..,N$ and a set of weighted connections $w_{ji} \in [-1,1]$, representing the strength and direction of influence from concept $C_i$ to concept $C_j$. In this study, each concept $C_i$ corresponds to a combination of a tabular feature $x'_{e,f}$ described by its associated fuzzy set $A_{e,f}$ . Each of the $N$ concepts $C_i, i = 1,..,N$, of FCM has a value $A_i = \mu^k_{e,j}\left(x'_{m,j}\right) \epsilon [0,1]$. Directed edges between concepts encode causal influence. There are three types of causal relationships: a) positive $w_{ji} > 0$ , which means that an increase in the value of $C_j$ causes an increase of the value of $C_i$, b) negative ($w_{ji} < 0$), indicating that increase in the value of $C_j$, causes a decrease of the value of $C_i$, and c) neutral $\left(w_{ji} = 0\right)$, meaning that there is no relationship between $C_j$ and $C_i$.

Directed edges in the FCM encode causally interpretable influence between features, and they are usually guided by the expertise of domain specialists. However, in this study the weights are estimated directly from data using fuzzy set interactions. To estimate this influence from data while preserving directionality, a fuzzy center-of-gravity (CoG) formulation is employed. Given the features $x'_{m,j}$ the influence of node $C_j$ to the node $C_i$ of FCM is defined as $b^k_{m,ji} = \min\left(\mu^k_{e,j}\left(x'_{m,j}\right), \mu^k_{e,i}\left(x'_{m,j}\right)\right)$ which captures the extent to which both features are linguistically active. The directed influence weight from node $C_j$ to $C_i$ is computed as:

$$w_{ji} = \frac{\sum_{x'_m \in D'_k}\left(x'_{m,j} \cdot b^k_{m,ji}\right)}{\sum_{x'_m \in D'_k}\left(b^k_{m,ji}\right)} \quad (3)$$

The resulting association scores are normalized to the interval $[-1,1]$ to comply with the standard formulation of FCM. This bounded representation allows weights to encode both reinforcing and inhibitory influences, facilitating (data-driven) causal interpretation and preventing unstable propagation during reasoning.

Optionally, domain knowledge or expert input may be incorporated by constraining the sign or existence of specific edges, yielding a structurally constrained FCM. Such constraints transform the learned FCM into a causally informed influence model, where edge directions and signs reflect assumed causal mechanisms rather than purely data-driven associations.

Once the FCM is constructed, a reasoning process is initiated and proceeds iteratively until convergence to a steady state. During this process, the activation value of each concept is updated according to:

$$A^{t+1}_i = f\left(A^t_i + \sum^N_{j=1,j\neq i} A^t_j \cdot w_{ji}\right) \quad (4)$$

where $A_i^{t+1}$ represents the value of $C_i$ at the iteration $t+1$, $w_{ji}$ is the influence of $C_j$ on $C_i$, and $f$ is a sigmoid function such as the log sigmoid, which maps the concept values within [0,1][17].

$$f(x) = \frac{1}{1+\exp(-x)} \quad (5)$$

The initial state vector $A^0 = \left[ \mu_{0,0}^{k}(x'_{m,0}), \dots, \mu_{E,d}^{k}(x'_{m,d}) \right]$ represents the initial concept values, for $t = 0$. This reasoning process enables the evaluation of IF-THEN scenarios and supports the generation of synthetic samples that are not only statistically consistent with the original data but also coherent with the assumed causal influence structure. The FCM reasoning step (Eq. 4) propagates the initial linguistic activation vector through the weight matrix for a fixed number of iterations or until a steady state is reached, adjusting each concept's activation according to the directed influence of all other concepts, thereby producing synthetic samples that reflect the inter-feature association structure captured during training.

### D. *Synthetic Sample Generation via Fuzzy Sampling and Causal Propagation*

Given the fuzzy distribution representation and the causally interpretable FCM constructed in the previous sections, synthetic samples are generated by jointly sampling linguistic feature states and enforcing causal consistency through fuzzy cognitive propagation. This process ensures that generated samples preserve both marginal feature characteristics and directed inter-feature influence relationships.

For each synthetic sample $\tilde{x} \in \mathbb{R}^d$ , generation begins by randomly initializing a linguistic activation vector in the fuzzy concept space rather than sampling numeric feature values directly. Specifically, an initial concept activation vector $A^0 = \left[ \mu_{0,0}^{k}, \dots, \mu_{e,f}^{k}, \dots, \mu_{E,d}^{k} \right]$ is generated by randomly sampling membership values from the fuzzy marginal distributions learned for each feature and class. These membership values represent the degree of activation of linguistic concepts and serve as the input to the FCM reasoning process. The randomly initialized linguistic state is then propagated through the FCM until convergence, according to the Eq.(4). This reasoning step enforces causal consistency among feature concepts, transforming the initially random linguistic configuration into a stable fuzzy state that respects the learned inter-feature influence structure.

Once a steady state is reached, the stabilized fuzzy activations define a coherent linguistic description of the synthetic sample. The final numerical feature vector is obtained through defuzzification. For each feature vector $\tilde{x}$ the fuzzy set (or sets) with the highest activation levels is identified, and a crisp value $\tilde{x}_f$ is randomly sampled from the support of the corresponding membership function. Sampling is biased by the membership degree, such that values closer to the fuzzy set prototype are more likely, while preserving variability across the feature domain. In cases where multiple fuzzy sets are simultaneously activated for a given feature, a weighted defuzzification strategy is applied, producing a crisp value as a weighted combination of the fuzzy set centers. This avoids hard discretization and ensures smooth transitions between linguistic concepts.

Through this process, synthetic samples are generated by transforming randomly initialized linguistic states into causally consistent fuzzy configurations, and subsequently into numerical feature vectors. This formulation highlights that data generation is driven by fuzzy reasoning and causal propagation, with defuzzification serving as the final step that materializes interpretable linguistic states into synthetic tabular data. Algorithm 1 highlights that synthetic sample generation is driven by fuzzy linguistic reasoning and causal propagation, with numerical feature values obtained at the final defuzzification stage.

## III. Experiments and Results

### A. *Experimental Setup*

The objective of the experimental evaluation is to assess the effectiveness of the proposed fuzzy distribution modeling with Fuzzy Cognitive Maps (FCM) for synthetic tabular data generation, and to compare its performance against established state-of-the-art baselines under comparable computational constraints [6].

All experiments were conducted exclusively on CPU, without GPU acceleration. This design choice reflects the intended scope of the proposed methodology as a computationally efficient alternative to resource-intensive deep generative models. Consequently, comparisons were restricted to widely adopted tabular data synthesizers that can be reasonably trained on CPU-based environments. More computationally expensive state-of-the-art models requiring extensive GPU resources were deliberately excluded to ensure fairness and reproducibility.

The proposed method was compared against three representative synthesizers from the Synthetic Data Vault (SDV) framework [18]: Gaussian Copula, a classical statistical baseline, CTGAN, an adversarial neural network for tabular data, and TVAE, a variational autoencoder-based model. All baseline models were used with their default SDV configurations. For each dataset, the number of generated synthetic samples was set equal to the size of the real training data.

The study used 3 real datasets from UCI repository for experimental evaluation[19]. The datasets were the Pima Indians Diabetes, the South African Heart Disease and the Statlog (German Credit). These datasets contain mixed continuous and categorical attributes and are commonly used in classification-oriented tabular learning tasks. Each dataset was split into training and test sets using a consistent partition across all methods. Synthetic data were generated solely from the training portion, while evaluation was always performed on the real test set. Normalization parameters, clustering prototypes, membership functions, and the FCM weight matrix are constructed exclusively from the training partition, the test set was not used during model construction or data generation.

### B. *Evaluation Metrics*

The evaluation followed a Train-on-Synthetic, Test-on-Real (TSTR) protocol, which is widely regarded as a reliable indicator of synthetic data utility. Three complementary categories of metrics were employed [18]: 1) Utility metrics to measure downstream task performance, a classification model was trained on synthetic data and evaluated on real test data. The reported metrics were accuracy,

**Algorithm 1** Synthetic Tabular Data Generation Using Fuzzy Distribution Space and FCM

**Input:** Fuzzy sets $A^k_{e,f}, y_{e,f} = k$, FCM weight matrix $W \in [-1,1]^{d \times E}$, Number of linguistic samples $M_s$, maximum iterations $T$, number of linguistics $e = 1, \dots, E$ , number of features per sample $d$
**Output:** Synthetic dataset $\widetilde{D^k} = \{(\tilde{x}_m, y_m = k)\}^{M_s}_{m=1}$., where k is the class
1: **For** m=1 to $M_s$ **do**:
2: //Step 1 Random linguistic initialization
3: Randomly initialize linguistic activation vector $A^0 = [\mu^k_{0,0}, \dots, \mu^k_{e,f}, \dots, \mu^k_{E,d}]$
4: //Step2 Causal reasoning via FCM
5: **For** t = 0 to $T$ **do**:
6: **For** each concept $C_i$ , $i = 1, \dots, d \times E$ **do**:
7: $A_i^{t+1} = f\left(A_i^t + \sum_{j=1, j\neq i}^{N} A_j^t \cdot w_{ji}\right)$
8: // Step 3: Defuzzification and numeric realization
9: **For** each feature $\tilde{x}_f, f = 1, \dots, d$ **do**:
10: Identify fuzzy sets $A^k_{e,f}$ with highest activation values
11: **if a single fuzzy set dominates then**
12: Sample $\tilde{x}_f$ from the support of $A^k_{e,f}$, biased by its membership function
13: **else**
14: Compute $\tilde{x}_f$ as a weighted combination of fuzzy set centers using their stabilized activation values
15: Store synthetic sample $\tilde{x}_m = [\tilde{x}_1, \dots, \tilde{x}_d]$ in $\widetilde{D}^k$

F1-Score and Area Under the ROC Curve (AUROC). 2) Fidelity metrics to evaluate how well synthetic data preserve the statistical properties of the original data, the following measures were used: Kolmogorov–Smirnov (KS) Complement, assessing marginal distribution similarity, Correlation Similarity, measuring preservation of inter-feature dependencies, Statistical Similarity, capturing aggregated moment-based statistics. 3) Privacy metrics were evaluated using Distance to Closest Record (DCR) Baseline Protection, which estimates the risk of record memorization and New Row Synthesis, verifying whether synthetic samples are exact replicas of real records. All metrics were computed using consistent evaluation pipelines across methods.

The quantitative results demonstrate that the proposed FCM-based approach achieves a high downstream utility, fidelity, and privacy across all evaluated datasets. Under the Train-on-Synthetic, Test-on-Real protocol, the method consistently outperforms CTGAN and remains competitive with TVAE and Gaussian Copula in terms of accuracy, F1-score, and AUROC. In particular, on the Statlog dataset, the proposed approach attains AUROC values nearly identical to TVAE despite operating with substantially lower computational complexity. Fidelity analysis further shows that the method preserves marginal distributions and inter-feature dependencies effectively, as reflected by high KS Complement, correlation similarity, and statistical similarity scores. The method achieves competitive KS Complement and maintains strong correlation and statistical similarity scores across all datasets. While Gaussian Copula achieves slightly higher fidelity on certain datasets, this often comes at the expense of reduced predictive utility, whereas the proposed framework maintains a balanced trade-off between distributional realism and task performance. Privacy evaluation indicates moderate to strong DCR baseline protection, consistently exceeding that of TVAE, while perfect new row synthesis confirms the absence of direct data replication. Perfect New Row Synthesis scores across all methods confirm the absence of memorization. DCR and New Row Synthesis assess memorization risk only, evaluation

TABLE I
TSTR UTILITY METRICS (CLASSIFICATION PERFORMANCE)

| Dataset | CTGAN | Gaussian Copula | TVAE | Proposed Method |
|---|---|---|---|---|
| **Accuracy** | | | | |
| Diabetes | 0.6623 | 0.7013 | 0.7338 | 0.7254 |
| SAHeart | 0.6022 | 0.6452 | 0.7312 | 0.7133 |
| Statlog | 0.6852 | 0.7593 | 0.8148 | 0.7769 |
| **F1-score** | | | | |
| Diabetes | 0.6434 | 0.6818 | 0.732 | 0.7022 |
| SAHeart | 0.5213 | 0.5834 | 0.6914 | 0.6381 |
| Statlog | 0.6405 | 0.7515 | 0.8161 | 0.7758 |
| **AUROC** | | | | |
| Diabetes | 0.6347 | 0.7021 | 0.8036 | 0.7566 |
| SAHeart | 0.4781 | 0.6191 | 0.7311 | 0.6911 |
| Statlog | 0.5563 | 0.868 | 0.9004 | 0.8989 |

TABLE II
FIDELITY METRICS (DISTRIBUTIONAL SIMILARITY)

| Dataset | CTGAN | Gaussian Copula | TVAE | Proposed Method |
|---|---|---|---|---|
| **KS Complement** | | | | |
| Diabetes | 0.7432 | 0.8344 | 0.8451 | 0.7951 |
| SAHeart | 0.7465 | 0.9039 | 0.819 | 0.8966 |
| Statlog | 0.8409 | 0.9428 | 0.8409 | 0.9001 |
| **Correlation Similarity** | | | | |
| Diabetes | 0.9044 | 0.9478 | 0.9532 | 0.9523 |
| SAHeart | 0.8895 | 0.9679 | 0.9432 | 0.8974 |
| Statlog | 0.9066 | 0.9391 | 0.9444 | 0.9356 |
| **Statistical Similarity** | | | | |
| Diabetes | 0.9157 | 0.9374 | 0.9573 | 0.9485 |
| SAHeart | 0.9159 | 0.9794 | 0.9383 | 0.9533 |
| Statlog | 0.9288 | 0.9805 | 0.9083 | 0.9214 |

TABLE III
PRIVACY METRICS

| Dataset | CTGAN | Gaussian Copula | TVAE | Proposed Method |
|---|---|---|---|---|
| **DCR Baseline Protection** | | | | |
| Diabetes | 0.4274 | 0.4422 | 0.2039 | 0.3971 |
| SAHeart | 0.4991 | 0.3421 | 0.2677 | 0.3533 |
| Statlog | 0.2320 | 0.1740 | 0.0662 | 0.2100 |
| **New Row Synthesis** | | | | |
| Diabetes | 1.0000 | 1.0000 | 1.0000 | 1.0000 |
| SAHeart | 1.0000 | 1.0000 | 1.0000 | 1.0000 |
| Statlog | 1.0000 | 1.0000 | 1.0000 | 1.0000 |

against membership inference and attribute disclosure attacks is left for future work.

Overall, these results highlight that the proposed fuzzy distribution–FCM framework delivers competitive synthetic data quality while operating under strict CPU-only constraints. Unlike deep generative models that rely on expensive neural optimization, the proposed approach exploits sparse linguistic activations and causal reasoning, yielding a lightweight yet expressive generative mechanism. Although the theoretical size of the FCM scales as $O(d \times E)^2$ with the number of features and linguistic values, only a small subset of concepts is active during reasoning and weak connections can be pruned to further reduce memory requirements, resulting in an efficient and scalable inference process. These findings demonstrate that high-quality synthetic tabular data generation can be achieved through interpretable, causally grounded fuzzy modeling without reliance on computationally intensive deep architectures.

To illustrate the interpretability of the proposed framework, consider a synthetic sample generated for the positive class of the Diabetes dataset. The corresponding linguistic activation vector prior to defuzzification may take the form $A^0$ = [High Body Mass Index, Medium Age, Low Diastolic Blood Pressure], representing a plausible patient profile. Following FCM propagation, the steady-state activations reflect inter-feature dependencies captured by the learned FCM. The final numeric values are then obtained through defuzzification from the dominant fuzzy sets. This intermediate linguistic representation provides an interpretable description of each generated sample, enabling inspection of the generation process, in contrast to black-box generative models.

## IV. Conclusion

This paper introduced a fuzzy distribution modeling framework for synthetic tabular data generation that reformulates the generation process as a causally informed fuzzy reasoning task rather than a conventional density estimation problem. By representing feature marginals through fuzzy sets and modeling inter-feature dependencies using sparse FCMs, the proposed approach enables interpretable, low-parameter, and computationally efficient synthetic data generation. Experimental results on multiple benchmark datasets demonstrate that the method achieves competitive utility, fidelity, and privacy compared to widely used state-of-the-art methods, while operating exclusively on CPU and avoiding the high computational cost of deep generative models. The incorporation of causal reasoning ensures coherence among generated features and supports transparent what-if analysis and linguistic interpretability. These findings establish fuzzy distribution modeling with FCMs as a principled and practical alternative for synthetic tabular data generation, particularly in resource-constrained, safety-critical, and decision-support applications. Future work will explore automated structure learning for causal graphs, adaptive linguistic granularity, and extensions to temporal and multi-relational tabular data.